\documentclass[10pt,a4paper]{article} 
\usepackage{lipsum}
\usepackage{url}
\usepackage{graphicx}
\usepackage{amsmath}
\usepackage[table]{xcolor}
\usepackage{array}
\usepackage[symbol]{footmisc}
\usepackage{natbib}
\setcitestyle{author-year}
\usepackage[nochapters]{./classicthesis} 

\begin{document}
    \pagestyle{plain}
    \title{\rmfamily\normalfont\spacedallcaps{Deep Reinforcement Learning for High Level Character Control}}
    \renewcommand*{\thefootnote}{\fnsymbol{footnote}}
    \author{\spacedlowsmallcaps{Caio Souza}\footnote{Correspondence: \textit{bycaioc2@gmail.com}} \footnotemark{} \setcounter{footnote}{1}, \spacedlowsmallcaps{Luiz Velho}\footnote{Instituto Nacional de Matem\'{a}tica Pura e Aplicada, Estrada Dona Castorina 110, 22460-320 Rio de Janeiro - RJ, Brazil.  }}
    
    \maketitle

    \begin{abstract}
        In this paper, we propose the use of traditional animations, heuristic behavior and reinforcement learning in the creation of intelligent characters  for computational media. The traditional animation and heuristic gives artistic control over the behavior while the reinforcement learning adds generalization. The use case presented is a dog character with a high-level controller in a 3D environment which is built around the desired behaviors to be learned, such as fetching an item.  As the development of the environment is the key for learning, further analysis is conducted of how to build those learning environments,  the effects of environment  and agent modeling choices, training procedures and generalization of the learned behavior. This analysis builds insight of the aforementioned factors and may serve as guide in the development of environments in general.
    \end{abstract}
    
    \tableofcontents

\renewcommand*{\thefootnote}{\arabic{footnote}}
\setcounter{footnote}{0}
    
 \section{Introduction}
    
Reinforcement learning have been a trending topic recently with the dissemination of Deep Learning techniques \citep{bengio2007scaling, lecun2015deep}. Various interesting applications have been made, from playing games like Atari \citep{mnih2013playing}, Go \citep{silver2017mastering} and StarCraft \citep{vinyals2019grandmaster},  to controlling actuators (muscles) on full body motion control simulation \citep{lee2019scalable} or a robotic hand solving Rubik's cube \citep{akkaya2019solving}, those are some examples in a vast range of applications. Here we want to experiment with high level controls of a character in a virtual 3D environment to do simple tasks like collect or fetch an object.

The chosen character is a dog named \emph{DogBot}. It has a set of animations (stand, walk, trot, run, jump, etc) and a heuristic traditional controller. Deep reinforcement learning is applied on it to select between the aforementioned actions at each instant $t_i$ to accomplish a desired task. This learned behavior can be then controlled by a higher heuristic or reactive pattern, which in its simplest form can enable or disable the fetch or any other behavior. 

\begin{figure}
\centering
\includegraphics[width=\textwidth]{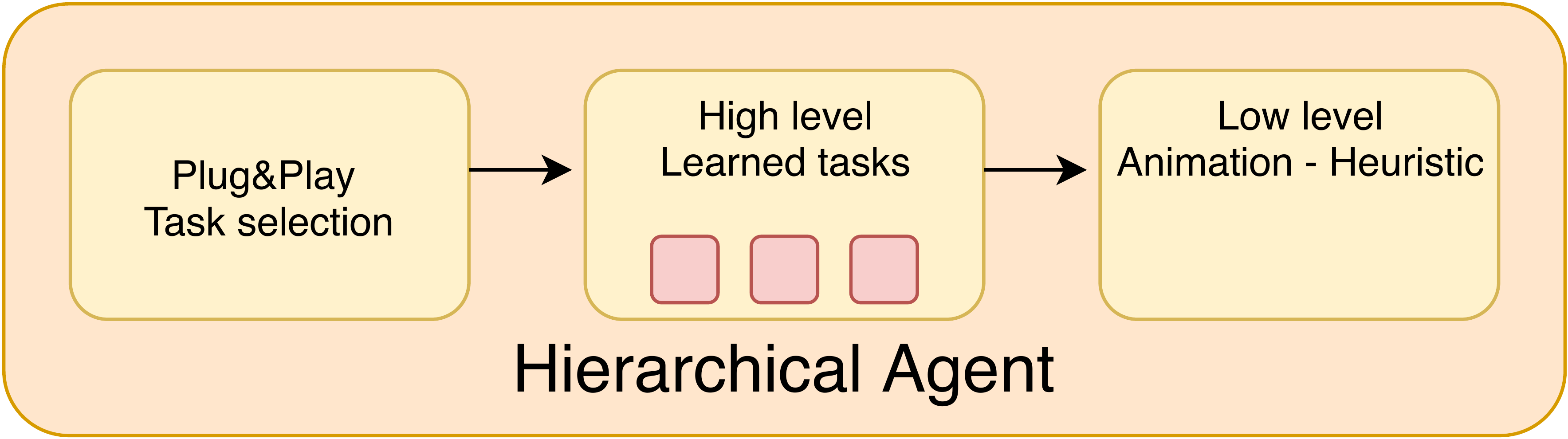}
\caption{Abstraction of the hierarchical agent proposed. The task selector can be either heuristic, or learned externally and independently, while the high level learned tasks are specific to a given low level controller which restricts how those tasks can be performed.}\label{fig:hier}
\end{figure}

This framework can be seen as an explicit hierarchical approach\footnote{\citep{barto2003recent} presents an overview of hierarchical reinforcement learning with the theory of semi-Markov Decision Processes, recent techniques and challenges of this field.} (figure \ref{fig:hier}), the lower level is the heuristic controller and animations, the mid level are the learned behaviors, each one comprising of a single task, and the final level abstracts the choice among tasks\footnote{One example of a higher level abstraction would be using command voice to select between tasks, it would be similar to products like Amazon Alexa and Microsoft Cortana, but working together with previous learned behaviors.}. This allows each level to be treated with some separation and freedom. One could use reinforcement learning in a end-to-end solution or mix heuristics with separated learned behaviors. 

\subsection{Motivation}

While there are plenty of successful cases of low-level training, developing such learning environments and characters demands a lot of additional work compared with the standard animation stack and also requires fine tuning. For such cases, achieving a desired artistic movement/animation can be challenging or not viable at all. Alternatively, current high level training  examples are mainly used to solve already posed problems with fixed rules, such as games from a player perspective. 

Our approach differs from those in its fundamental principle: it does not solve an existing game, but creates it own game environment and rules to learn a desired task. This gives the freedom of balancing between the learning difficulty, which accounts for the generalization, and the control over the actions which are the heuristics and animations. All that with access to the underlying environment model\footnote{Access to the environment model gives the possibility of shaping the reward process easily and possibly avoiding or softening one of the fundamentals problem of reinforcement learning: The credit assignment\citep{sutton2018reinforcement}.}.

The mixed use reinforcement learning and traditional animation simplifies the artistic control over the final result. It gathers the best of each world: the traditional animation and heuristic gives control over the set of actions whereas the reinforcement learning gives the possibility of the agent generalizing for different situations.

\subsection{Organization}

This paper is organized as follow: section \ref{sec:modeling} contains the modeling of the environment used and properties for environments in general. The section \ref{sec:training} contains the training methodology and parameters. Next, section \ref{sec:expr} presents the results and its discussion. Finally, we conclude with section \ref{sec:conclusion}.

\section{Environment and Agent Modeling}\label{sec:modeling}

The modeling of the virtual environment and agent were made with Unity Editor 2019.3 using its physics engine simulation.
\subsection{Environment}
\begin{figure}
\centering
\includegraphics[width=\textwidth]{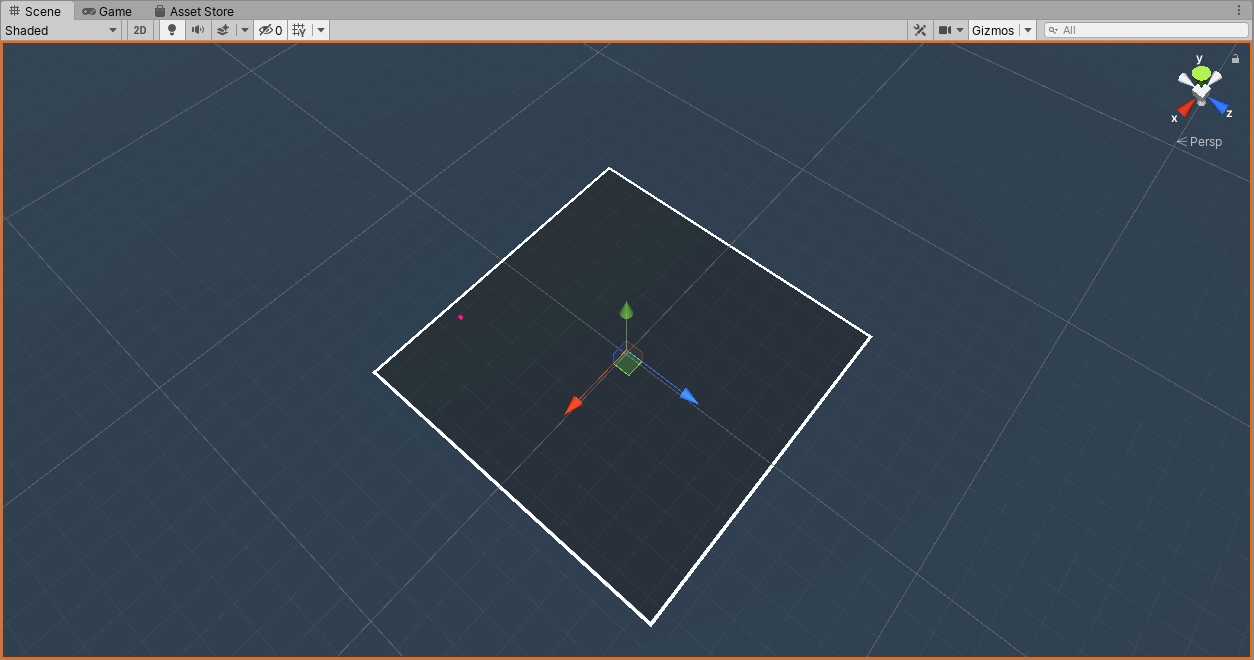}

\caption{Top View from environment inside the Unity Editor.}\label{fig:env}
\end{figure}

The environment is a Unity scene where the character can act and interact with objects. It consists of a square gray plane $110\times 110$m with a white border of $1$m diameter, which visually delimits the area of interest. While it is possible to walk past this area, if the agent go past it, a final state (game over) is reached leading to a reset. The figure \ref{fig:env} shows the top view of the environment in the Unity editor.

\begin{figure}
\centering
\includegraphics[width=\textwidth]{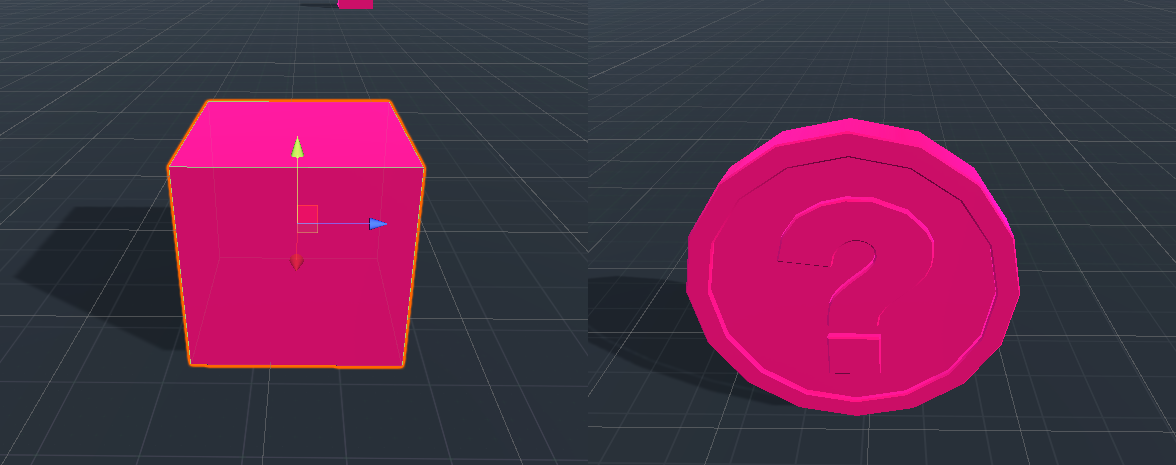}
\caption{Objects which can be collected inside the training environment. They are purposefully big and color differentiated for training purposes. This is an advantage of virtual environments, where what is seen by the agent does not necessarily is seen by a player. A simple visualization can be used for the agent, while a complex and entirely different for a player.}\label{fig:objs}
\end{figure}

Inside the delimited area the following objects can be found (figure \ref{fig:objs}):
\begin{itemize}
\item{Collectibles}
	\begin{itemize}
		\item{(Simple Geometry) Cube with $1$m edge}
		\item{(Complex Geometry)  Coin with $1.5$m diameter}
	\end{itemize}
\end{itemize}

The choice of objects is arbitrary, but this is an advantage of virtual environments: What is seen by the agent don't need to be the same which is seen by a player, hence the freedom of using simple objects and rendering for the agent, albeit a complex scene may be displayed to the player. Another important point is the size of the objects, with later usage of visual information at low resolution ($84\times 84$ pixels) rises the need of big objects.

\subsection*{Environment Observability}

An important conceptual distinction of environments is whether it is completely or partially observable. Although it's described as an environment property here, it is directly related to the agent's perception of its world.

\begin{itemize}
\item{Completely Observable - All information of the environment state is observable.}
\item{Partially Observable - Some of the information of the environment state is observable.}
\end{itemize}

One example of a completely observable environment is:\\
Assuming the previous environment area without obstacles and with a single collectible object as goal. If the observations are set as the agent position and direction, the direction to the target, the target position and the distance and position of the border. It is a completely observable configuration in the sense that independent of the agent state (position and direction) it always have complete information of itself and the target to complete its goal\footnote{``All information'' is relative to the important information to complete a given task. While it is not possible to draw the complete scene from the given observations, it is sufficient to complete the task at any state/time without the need for information such as object shape, color, etc. For this simple case is possible to define if it is completely observable, but complex environments may not be trivial to assert such definition.}. Indeed one could even write a heuristic to solve this task.

In contrast, following the same example, if only visual information were used, (i.e, a 2D image from the agent's vision cone), depending of the agent position it may or may not ``see'' the target, neither know its exact position. In this case the sensing of the agent varies depending of its state and it is always a partial view of the environment which may or not contain relevant information.

 This differentiation of how the environment is perceived and how much information is available per observation is an important component of learning performance.

\subsection{Task}

The agent's goal can be put as a general mobility task, given a stimulus it needs to reach the target point. Inside this general task two specialized tasks are implemented:

\begin{itemize}
\item Collecting an object:
reach the object position, when the agent collides with it the object is removed from the scene, i.e., collected.
\item Fetch:
reach the object and go back to its initial position. It also can be thought as reaching two objects, for example, the stick and then who threw it.
\end{itemize}

While these can all be cast as essentially the same task, their difference comes of how they are modeled inside the virtual environment and which information is available to the agent to complete them. That said, they are practical examples of mobility tasks.

\subsection{Agent}

\begin{figure}
\centering
\includegraphics[width=\textwidth]{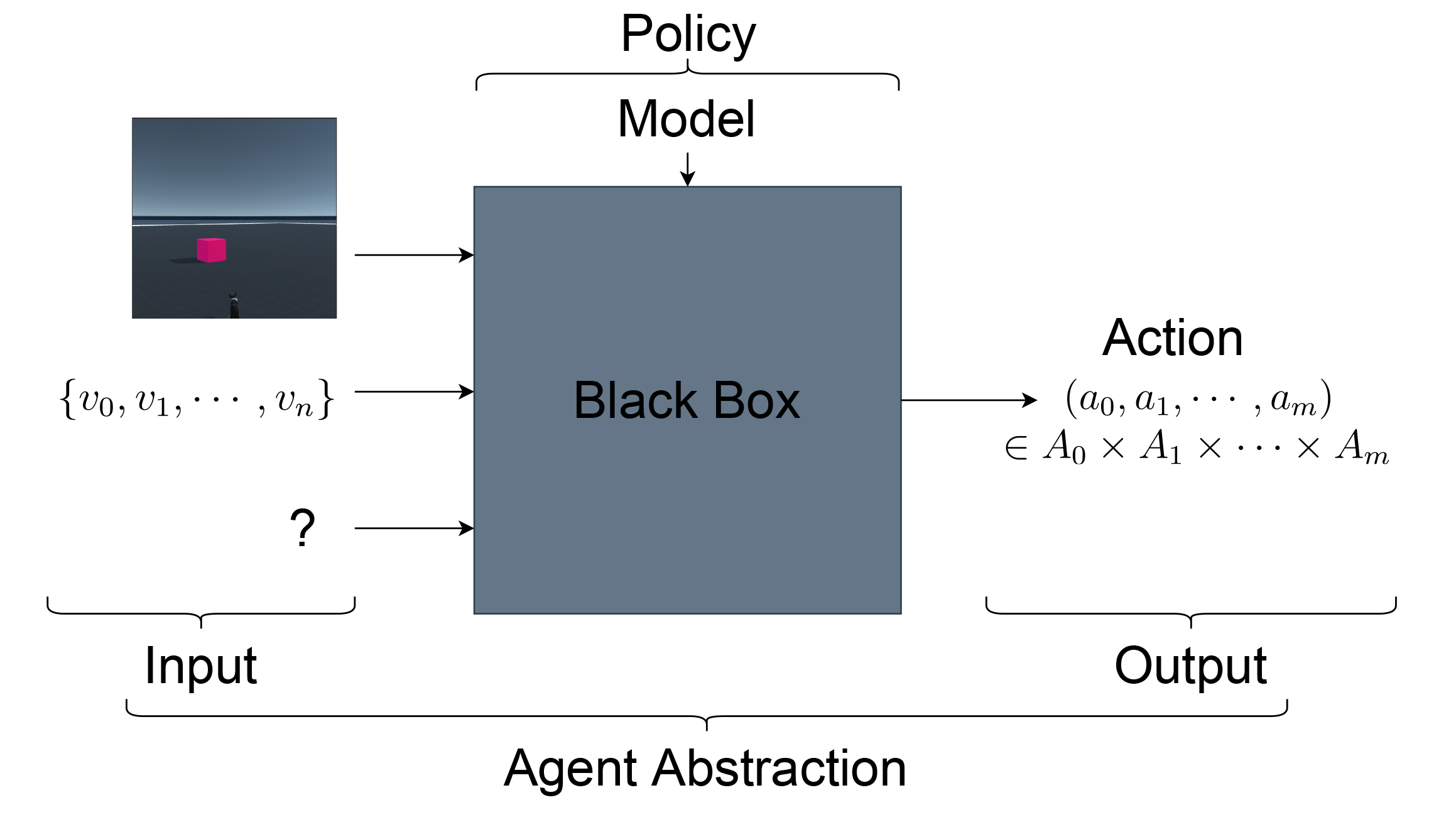}
\caption{Agent abstraction as a black box with input/output. Any kind of input can be used, they are processed by policy and output a pre-defined set of actions.}\label{fig:agent-abst}
\end{figure}

The agent is an entity abstraction which is itself a \emph{behavior policy}. Nevertheless, when referred without a specific policy it can be imagined as an entity with sensors collecting observations and actuators interacting, where a policy can be plugged-in linking an observation with a given action\footnote{In programming it would be equivalent to the concept of an interface which define empty methods with input and output, where the implementation itself would be the policy.} as show in the figure \ref{fig:agent-abst}. 

The important points in a agent are: what it observes to take actions, how both the observations and actions are encoded and how the associated rewards are distributed, these topics are further discussed in the following sections.

\subsection*{Character Controller}

\begin{figure}
\centering
\includegraphics[width=0.5\textwidth]{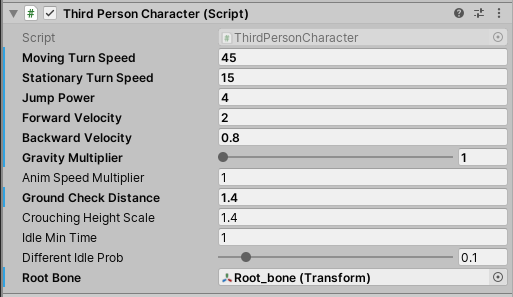}
\caption{Character intrinsic properties inside Unity Editor. Those properties controls how the animations are performed and are not visible to the policy or the agent itself.}\label{fig:char-ctrl}
\end{figure}

The Unity character controller is the lower level hierarchy of agent's actuators. It can be considered intrinsic to the agent and it controls the agent's velocity, turn speed, gravity and other effects of the character physical properties. The detailed list of the parameters used in the training are in the section \ref{sec:config}, while the figure \ref{fig:char-ctrl} shows its interface inside the Unity Editor.

\subsection*{Agent Observation}

An observation is any kind of sensing made from the agent itself or the environment state. Those need to be encoded by numbers that serves as input to the behavior policy.
Here, two types of observations are employed:

\begin{itemize}
\item Vector observation: complete observable, composed of hand-crafted features:
	\begin{itemize}
	\item Normalized direction to target: $ d_{\text{target}} = (x,y,z)$,  $\lVert d_{\text{target}} \rVert_{2} = 1$
	\item Normalized distance to border: $d_{\text{border}} = (x,y)$,
	 $\lVert d_{\text{border}} \rVert_{\infty} < 1 \rightarrow \text{inside}$, 
	 $\lVert d_{\text{border}}\rVert_{\infty} \geq 1 \rightarrow \text{outside}$
	\item Linear velocity: $ v_{\text{linear}} = (x,y,z) \text{m/s}$
    \item Angular velocity: $ v_{\text{angular}} = (x, y, z) \text{rad/s}$
	\item Normalized agent forward direction: $ d_{\text{forward}} = (x, y, z)$,
	 $\lVert d_{\text{forward}} \rVert_{2} = 1$
	\item Normalized agent up direction: $ d_{\text{up}} = (x, y, z)$,
	  $\lVert d_{\text{up}} \rVert_{2} = 1$
	\item Agent local position (agent's position referent to the center of the environment): $ p_{\text{local}} = (x,y,z) $
	\end{itemize}
\item Visual observation: partially observable, down-sampled from the original rendered image, 3rd-person-like camera aligned with agent forward direction:
	\begin{itemize}
	\item 2D image:  matrix $I_{84\times 84 \times 3}(r,g,b)$\footnote{DogBot's experiment uses RGB images, but these virtual environments allows the use, for example, of depth maps or any other 2D input.}
	\end{itemize}
\end{itemize}

\begin{figure}[ht!]
\centering
\includegraphics[width=\textwidth]{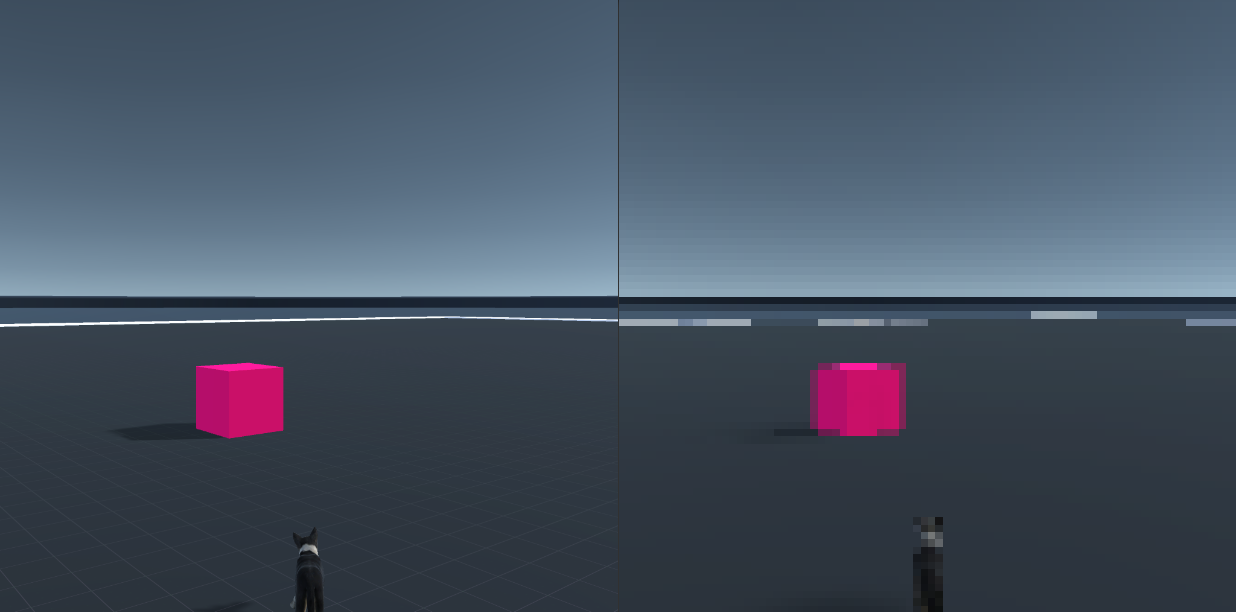}
\caption{(Left)Agent's third-person camera. (Right)Downsampled visual observation. The size of agent's input is direct related to the computational complexity when training, hence down-sampling is used to balance this cost.}\label{fig:visual-obs}
\end{figure}

The first encoding takes a total of 20 numbers (floats) as observation. These were calculated from the agent/environment state and are much like as the observations taken by Puppo, the Corgi\footnote{Puppo is a demo from Unity 3D where a dog uses a learned behavior acting direct in its joints (low level control). We cite it here, because it is used for later comparisons.}\citep{puppo}, which is a demo made by Unity. However there are some changes to fit our modeling. First, the Normalized distance to the border is added, because it is relevant in an unbounded environment to keep the agent inside the desired area. Last all the joint angles and torque information were removed. The Unity Puppo works on a low-level control of the joint angles and torque while this new agent works on a high-level control with animations. Those differences are related to how the agent senses itself, while how it senses the rest of the environment and its target remained the same.

Next, the visual encoding consists of 84×84×3=21168 numbers (floats), in a 2D RGB image. This image is the direct rendering of the agent's view (figure \ref{fig:visual-obs}). This kind of observation is interesting for many applications, despite being much more complex and less complete than the first encoding in 20 numbers, because only partial information can be inferred from it, it can be more general in some aspects. For example if a variable number of collectibles is admitted in the environment, the first encoding would not be able to handle that variation, but using a 2D image the agent could deal with many objects (because the image is already a partial view of the environment).

\subsection*{Observation Constraint}

One differentiation about those observations is with respect to its constraints. They can be:
\begin{itemize}
\item Self-contained - Depends only on the agent state and sensors.
\item Environment-model dependent - Depends on the environment underlying mechanics.
\end{itemize}

With those criteria, the 2D image is an agent self-contained sensing while the vector observation needs access of the underlying environment model to be processed and fed to the agent (i.e. target position). This is directly related to the applicability and generalization of the agent. 
The disadvantage of a model-dependent agent is: it cannot be placed in a different environment where it does not have access to the underlying model.

\subsection*{Agent Action}

The DogBot's actuator(s) is a character inside Unity. It is composed of various animations and a controller (with blend trees and alikes) which receives four parameters controlling the X-Axis velocity, the Y-Axis rotation and booleans jump/crouch.

For the actions encoding two schemes were used continuous and discrete:

\begin{itemize}
\item Continuous action space:
	\begin{itemize}
	\item Forward and backward movement: $\in[-1,1]$
	\item Steering left and right: $\in[-1,1]$
	\item Jump: $j\in[-1,1]$,$j>j_0$
	\item Crouch: $c\in[-1,1]$, $c>c_0$
	\end{itemize}
\end{itemize}
where $c_0$ and $j_0$ are threshold parameters intrinsic to the agent controller.
\begin{itemize}
\item Discrete action space:
	\begin{itemize}
	\item Forward and backward movement: $\{\text{backward, none, walk, trot, run}\}$
	\item Steering: $\{\text{left, none, right}\}$
	\item Jump: $\{\text{true, false}\}$
	\item Crouch: $\{\text{true, false}\}$
	\end{itemize}
\end{itemize}

Each item is an action branch, and can be choose simultaneously. For the case of Jumping/Crouching, despite being separated branches, priority is given to Jump action over the Crouch as it is not physically possible to do both at the same time. While the encoding for actions are arbitrary, they reflect the ranges of the Unity character controller such as max velocity and turn speed, which are configurable but from the agent's perspective are intrinsic to its actuator.

\subsection{Reward}

The entire universe of reinforcement learning is based on encouraging the best behavior through rewards (much like it is done when teaching a trick to a pet), in other words, rewarding good actions according. It can also be posed as a optimization problem of maximizing the total reward \citep{sutton2018reinforcement}.

In this scenario two perspectives are brought up: developing a good reward signal is the key to being able to solve this optimization problem, yet most of times it is not as easy to qualify a given action and state pair individually, but only the final outcome of a sequence of actions and states. In theory, even for the cases where only the final outcome is rewarded, in the limit after many (infinite) experiences it would be possible to learn an optimal behavior policy. Sadly, in the real world limited resources are available, hence the art in learning good policies lies in modeling good environments, rewards and algorithms as much as possible\footnote{Here good performance means reduced time and sample complexity.}.

The DogBot agent experiments with two types of reward:
\begin{itemize}
\item Per action reward:
	\begin{itemize}
	\item Positive reward is assigned if the agent gets closer to the target, negative reward is assigned if it distances itself from it. The formula used in this case is \(r = +0.01 (v_{\text{linear}} \cdot d_{\text{target}})\) or $+1$ if the agent achieved the objective.
	\end{itemize}
\item Sparse reward: 
	\begin{itemize}
	\item Positive reward \(+1.0\) is given when the agent reach its destination, i.e., the collectible.
	\end{itemize}
\end{itemize}
In all cases, leaving the training area leads to a negative reward of \(-1.0\) ending the episode. Also a small negative reward \(-0.0005\) is given at each time step \(t_i\). It was chosen to be close to \(-\frac{1}{\text{\# steps}}\) so the accumulated penalty would not saturate the total reward signal. This is widely used as a time penalty to stimulate the completion of a task in the shortest time.

It's also important to classify these rewards in another way: The first reward needs a broader knowledge of the environment to be calculated depending both of the agent and environment underlying state. Conversely, the second could be completely assigned with the agent's sensing, but brings in the credit assignment problem which can be translated as finding \textit{how much each previous state contributed to the present reward?}

This differentiation is interesting because it resembles the agent self-contained observation property and have implications in the learning performance. In this case, even for the model dependent reward, after the learning process it does not prevent the agent to be used in a new unknown environment.

\section{Training}\label{sec:training}
  
The tools used for training were Unity Editor 2019.3 and its machine learning framework ML-Agents \citep{juliani2018unity} in its version v0.13.1. All experiments use the Proximal Policy Optimization \citep{schulman2017proximal} algorithm. Experiments were run only once given that they are computationally demanding. While the results are expected to be similar between runs, when analyzing the results some variability should be taken in account.

The ML-Agents framework exposes some parameters related to it's built-in models (network architecture), they will be specified inside the section \ref{sec:config}. Other specific details of the network itself are omitted because they are the default of ML-Agents and can be found in their online documentation \citep{juliani2018unity}. This choice was made because the focus is on the development of the agent, environment and reward signal instead of network architecture.

The training of the ``Unity Puppo, the Corgi" uses the same configuration presented in the table \ref{tab:train}. The environment is the same provided by \citep{puppo}, ported to the newer version of ML-Agents(v0.13.1) used here. The only change to the original environment was its training area size to match DogBot's area size.
 
\subsection{Configuration}\label{sec:config}

The parameters used both for the character controller and and ML-Agents are presented in the tables \ref{tab:ctrl} and \ref{tab:train} respectively.

\begin{table}
\caption{DogBot's character controller parameters.}\label{tab:ctrl}
\begin{tabular}{p{0.2\linewidth}p{0.15\linewidth}p{0.55\linewidth}}
\hline
\textbf{Name} & \textbf{Value} & \textbf{Description} \\ 
\hline 
Moving turn speed & $45.0$ deg/s & Turn speed when not stationary\\
\hline
Stationary turn speed & $30.0$ deg/s & Turn speed when stationary\\
\hline
Jump power & $5.0$ m/s & Vertical velocity applied when jumping\\
\hline
Forward Velocity & $9.0$ m/s & Maximum forward velocity\\
\hline
Backward Velocity & $2.0$ m/s & Maximum backward velocity\\
\hline
Gravity multiplier & $1.0$ & Multiplier for gravity simulation\\
\hline
Anim speed multiplier & $1.0$ & Multiplier for animation time scale\\
\hline
\end{tabular}
\end{table}

\begin{table}
\caption{ML-Agents training parameters.}\label{tab:train}
\begin{tabular}{p{0.225\linewidth}p{0.125\linewidth}p{0.55\linewidth}}
\hline 
\textbf{Name} & \textbf{Value} & \textbf{Description} \\ 
\hline 
batch size (Continuous) & $4096$ & Number of samples used for each optimization step for continuous action space.\\
\hline
batch size (Discrete) & $256$ & Number of samples used for each optimization step for discrete action space.\\
\hline
buffer size & $40960$ & Number of samples collected for each policy update.\\
\hline
hidden units & $512$ & Number of neurons per hidden layer.\\
\hline
num layers & $2$ & Number of hidden layers used for the model.\\
\hline
learning rate & $3.0 \times 10^{-4}$ & Initial learning rate for training.\\
\hline
max steps & $2 \times 10^7$ m/s & Number of total simulation steps (actions) taken for training.\\
\hline
num epochs & $5$ & Number of times each collected observation is used for training.\\
\hline
time horizon & $1000$ & Horizon for learning, it represents how far in time steps one action can influence a past reward.\\
\hline
gamma & $0.995$ & Discount factor, it represents how much of a n-future reward ($R_n$) is assigned to a present action in the form $\gamma^nR_n$.\\
\hline
Curiosity strength & $0.1$ & Strength of the curiosity intrinsic reward signal.\\
\hline
Curiosity gamma & $0.99$ & Discount factor for the curiosity reward.\\
\hline
Visual encoding type & nature\_cnn & Type of architecture used for the convolutional layers for the visual observation encoding.\\
\hline
Max episode length & 5000 & Maximum number of steps until an episode ends.\\
\hline
\end{tabular}
\end{table}

\section{Results}\label{sec:expr}

The table \ref{tab:exp1} and \ref{tab:exp2} contains the result of various trained models evaluated on the test scene containing $n$ collectibles which randomly re-spawn when collected. The evaluation metric is the Score, (the number of objects collected over all episodes) and Reset, (the number of times the agent was reseted due to leaving the training area). It ran for $200$ episodes or $10^6$ steps.

The training convergence of the experiments are presented in the section \ref{sec:convergence} followed by the analysis and discussion of the results in the section \ref{sec:discussion}.

\begin{table}
\caption{Results for the various models on the standardized test scene. In order the columns are: Experiment number(Exp), observation type(Obs. Type), action type(Act. Type), number of collectibles in the training  environment(Train Env.), Reward Type(Reward Type), number of collectibles in the test environment (Test Env.), Score(Score) and Reset(Reset).}\label{tab:exp1}
\resizebox{\columnwidth}{!}{
\begin{tabular}{cccccccc}
\hline
\textbf{Exp} & \textbf{Obs. Type} & \textbf{Act. Type} & \textbf{Train Env.} & \textbf{Reward Type} & \textbf{Test Env.} & \textbf{Score} & \textbf{Reset}\\
\hline
1 & Vector & Discrete & 1, box & Per action & 1, box & 1027 & 61\\
\hline
2 & Vector & Continuous & 1, box & Per action & 1, box & 2324 & 82\\
\hline
3 & Vector & Discrete & 1, box & Sparse & 1, box & 4 & 670\\
\hline
4 & Vector & Continuous & 1, box & Sparse & 1, box & 0 & 0\\
\hline
5 & Visual & Discrete & 1, box & Per action & 1, box & 1370 & 7\\
\hline
6 & Visual & Continuous & 1, box & Per action & 1, box & 2883 & 0\\
\hline
7 & Visual & Continuous & 24, boxes & Sparse & 1, boxes & 12 & 0\\
\hline
8 & Visual & Continuous & 24, boxes & Sparse & 24, boxes & 7119 & 3\\
\hline
9 & Visual & Continuous & 1, box & Per action & 24, boxes & 6163 & 0\\
\hline 
\end{tabular}}
\end{table}

\begin{table}
\caption{Results for the down-scaled environment on the standardized test scene. In order the columns are: Experiment number(Exp), observation type(Obs. Type), action type(Act. Type), number of collectibles in the training  environment(Train Env.), Reward Type(Reward Type), number of collectibles in the test environment (Test Env.), Score(Score) and Reset(Reset).}\label{tab:exp2}
\resizebox{\columnwidth}{!}{
\begin{tabular}{cccccccc}
\hline
\textbf{Exp} & \textbf{Obs. Type} & \textbf{Act. Type} & \textbf{Train Env.} & \textbf{Reward Type} & \textbf{Test Env.} & \textbf{Score} & \textbf{Reset}\\
\hline
10 & Visual & Continuous & 1, box & Per action & 1, box & 4606 & 599\\
\hline
11 & Visual & Continuous & 24, boxes & Sparse & 1, box & 1502 & 241\\
\hline 
\end{tabular}}
\end{table}
\subsection{Training Convergence}\label{sec:convergence}

\begin{figure}[ht!]
\centering
\includegraphics[width=\textwidth]{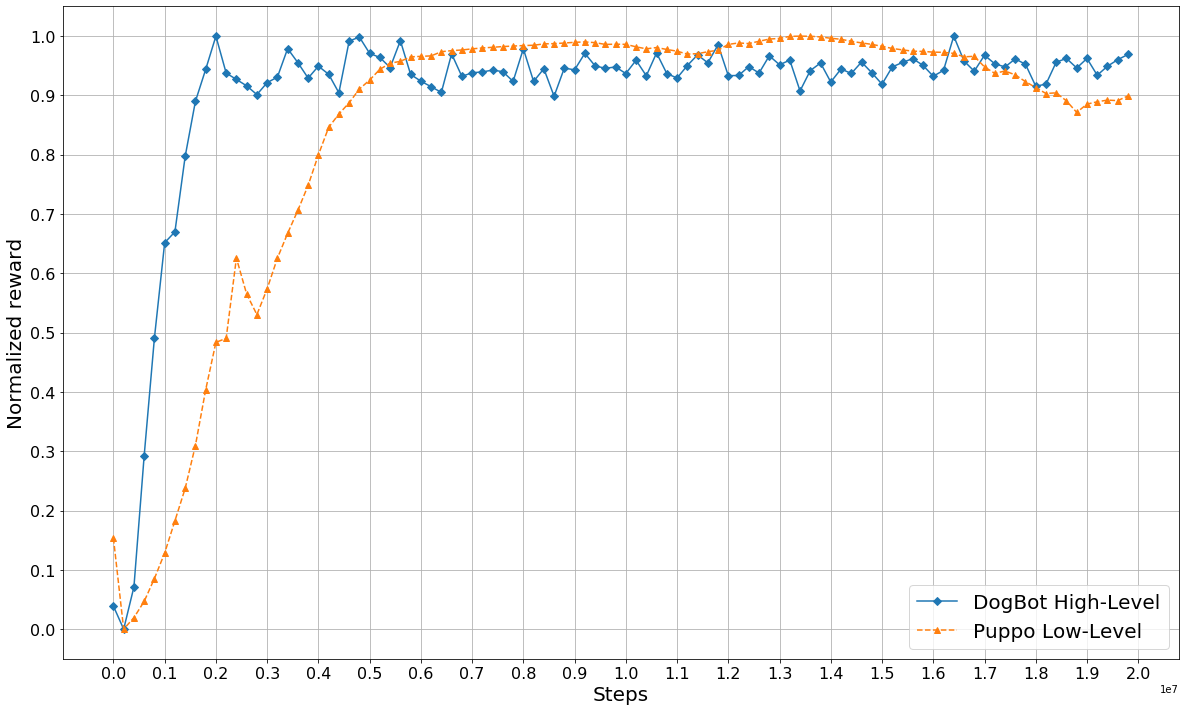}
\caption{Comparison of the training convergence of the ``Puppo, The Corgi" which uses a low-level control approach versus the DogBot high-level character control.}\label{fig:convg}
\end{figure}

The figure \ref{fig:convg} presents the comparison of Puppo and DogBot training convergence. This result is not directly comparable with the following figure \ref{fig:convg2}, because the modeling\footnote{Here the modeling accounts for observation, reward, and final state, besides the changes to fit the high level type control proposed.} used is the the same from Puppo, for comparison purposes, which differs ours.

The figure \ref{fig:convg2} shows the evolution of the training from various configurations. This figure intends to show the convergence and evolution of the training procedure, while the previous tables (\ref{tab:exp1} and \ref{tab:exp2}) presents specific result for each configuration in a standardized test scene.

\begin{figure}[ht!]
\centering
\includegraphics[width=\textwidth]{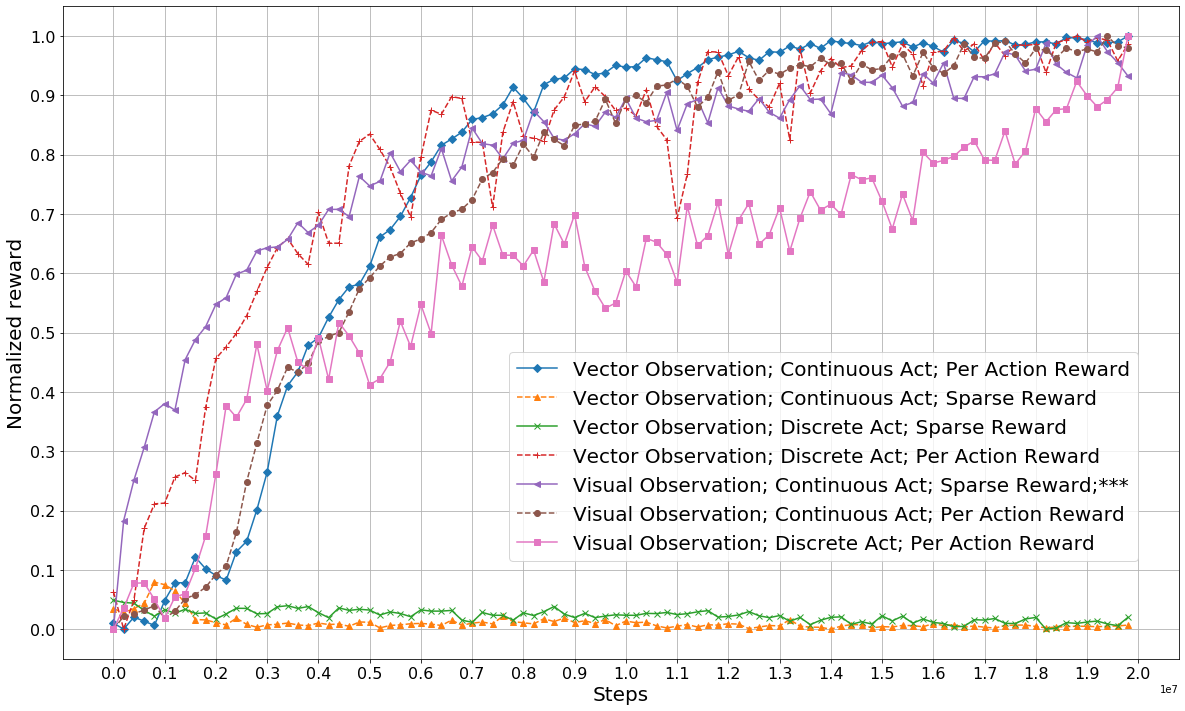}
\caption{Normalized reward evolution during the training process. The naming scheme is the following: ``\textit{observation type}"; ``\textit{action type}";``\textit{reward type}". The entry marked with \emph{***} stands for the environment with multiple collectibles, while all others were using a single collectible.}\label{fig:convg2}
\end{figure}

\subsection{Discussion}\label{sec:discussion}

\subsection*{High and low level control}

Low level control gives the possibility of an end-to-end learning, but in the present case it converges slower than high level learning. Another downside is that it may requires more work for developing such agents and fine tuning to specific actions. For virtual scenarios and entertainment tasks it may be more convenient to use traditional animations, which are widely available and offers easier control over the final result.

\subsection*{Action branches and time interval}

One difficulty when training with all action branches active at the same time was that it would not learn anything. Starting from a random policy, the agent would be stuck alternating between forward, backward, left and right. This problem with the axis movement is probably due to the interval between the actions being short and not allowing the agent to commit to a give action for enough time\footnote{This is a direct limitation of the character controller which smooths the transition between animations, and in our case, the time between transitions were greater than the agent's decision interval.} until a strong reward signal were obtained. 

Nevertheless increasing the action interval would lead to unwanted delays in the reaction time of the agent. Four solutions which worked well are:

\begin{itemize}
\item Giving a slightly reward for moving forward.
\item Adding a bias in the actions favoring the forward movement.
\item Training the action's branch separated.
\item Restricting the Y-Axis to forward movements only.
\end{itemize}

\subsection*{Observation type}

Visual observation achieved better results than vector observation when using the same reward system. It also could learn faster in the environment with many collectibles, where vector observation can't be applied. Despite it requiring more computational power, it shows that passing raw inputs may achieve better performance than having hand crafted features.

\subsection*{Action space}
Continuous actions space achieved better results than the discrete action space, yet its convergence was slower. One point to notice is the batch size for continuous and discrete action space were different, and they were not extensively searched for the best result. The use of smaller batch size for discrete actions space is justified because the larger the batch size more samples are averaged in the training process and the in-between or average of discrete or categorical actions may not make sense at all, hence it is advised to use small batch sizes for these cases. On the other hand, for continuous actions the average between samples makes sense or exists\footnote{Another way to think about it is how their sampling is done.
For discrete space each action has a given probability, which are disjoint, the classes itself are not related to the \emph{encoding} or numbers used to represent it.
For continuous space the final value is sampled from a distribution with a given mean and variance, hence the need to approximate it from various sample steps obtained in the training process.}.

\subsection*{Reward}
Using a per-action reward strategy worked in all cases, while sparse reward only worked in the scene with multiple objects. It is clear that even with the drawback of the agent possibly exploiting the reward instead of learning the true objective a per-action strategy is more reliable. 

Another crucial point is that the per-action reward developed a good exploration policy when the agent was not seeing the collectible. In these cases it ran to the opposite side of the training area, while the agent trained with sparse reward usually get stuck running in circles until it is close enough to notice the collectible. It explicits how having access to the underlying environment when developing the reward system can be beneficial.

The sparse reward, given only when an object is collected, can't be exploited, but makes the learning process entirely dependent of the environment's difficulty. For those cases a good approach is using a curriculum\footnote{Curriculum learning is an approach of increasing the environment difficulty according with the agent learning which was proposed in \citep{bengio2009curriculum}}, starting with very easy tasks and increasing the difficulty as the agent performance increases. An example of such curriculum could be starting the environment with many objects or small area and decreasing/increasing it according as the agent learns.

\subsection*{Multiplicity of collectibles}

Having multiple objects in the scene completely changed the result of using sparse reward, from unable to learn anything to achieving the best result in the test environment with multiple collectibles. 

Giving the agent an environment where it can often achieve a goal even with a random policy is the key point to use sparse reward. It is a good strategy to be used with curriculum learning, in this case, the difficulty level could be easily controlled by the amount of collectibles. While it was not tested here, one can believe that an agent trained with such curriculum could perform better especially in the test case where it performed bad with a single collectible.

\subsection*{Environment variation}
Two variations where tested, changing the size of the environment and the number of collectibles. Those variations shown interesting behaviors.

First the agent trained with multiple collectibles, (which was not able to perform well in the single collectible environment), had a good performance when the size of the environment was reduced. From visual inspection, what was happening was: the agent had a short sight and this effect was attenuated by testing on a smaller area.

Then, the agent trained with a single object performed very well with multiple collectibles, but not as good as the one trained with many objects. Also from visual inspection, the agent would only turn to the right, which indeed works but is not the optimal behavior with multiple objects. In the case of a single object, which most of times was far away, it would not change much, but in the case of multiple objects it was a limiting factor.

These behaviors reinforces the idea of the need for variation in the training environment. It also shows how details of the environment can introduce undesirable side effects or completely ruins the learning of the desired behavior.

\subsection*{Learning generalization}
Although the agent could somewhat generalize well to multiple and single objects it wasn't as good as the performance of the trained environment. When running on a smaller area it could not cope with the delimited marks and the agent left the area much more than when using the original size. One possible cause is the intrinsics of the character controller which does not allow the agent to turn fast and was accentuated by the decrease in the training area.

For achieving good generalization it must experience great variation of the environment and goal, which was only partially provided. Yet, this exposition need to be done without a steep difficulty variation, or the agent may not be able to learn anything. 

\subsection*{Other considerations}

Some factors affecting the agents behavior were covered by the agent and environment design, yet it is far from being extensive. Other than that, factors such as network architecture and algorithms were not evaluated. Using architectures with memory and different algorithms could have influenced in both the convergence and quality of the learned policies. An example of a possibly improvement from using architectures with memory is for the case where the agent run in circles, it is doable to look around when the agent can't see the goal, but doing it repeatedly could be avoided if memory of past states where present.

\section{Conclusion}\label{sec:conclusion}

One of the key features to develop is certainly the reward system. Making the agent able to have feedback even with a completely random behavior is a must, be it through a per action reward or a easier environment where rewards are not much sparse. Even so, it may get stuck if the action space is complex or if there is not enough time to commit to each individual action. This was exemplified by the lack of convergence when using the full action space despite having a per action reward. Nevertheless, with sparse reward and increased amount of objects in the scene leaded to convergence.

The type of observation used also shown some less intuitive behavior: while the vector observation had all the data needed to heuristically solve the problem, the visual observation had a better result. It may be due to encoding which is compact and human understandable, but may not be the best for learning, since it relies on common human knowledge and concepts. Letting the model extract the features from visual observation is computationally expensive for training, but it achieved better policies and is easily extensible to new environments.

Modeling and training agents to complete tasks, (in the way one would expect a human being to do so), is a complex problem. Good policies were learned, although side effects and not so much intuitive results were obtained too. This explicits how an agent can exploit unnoticed details of its environment in unpredictable ways for good, bad and ugly things. Although the agent may solve the given task, expecting human like (in our case dog like) behavior may mislead the modeling and comprehension of the results.

Time and sample complexity are common metrics for evaluating performance, but human resources should also be taken in account. Developing good environments demands a good amount of human resources which can be more expensive than computational resources. This alone can motivate the research for general good practices and guides.

Here, nor the neural network architecture/parameters neither the environment graphics complexity were evaluated, but the modeling of the environment, agent and reward. Certainly these aspects are important, but much has been done in those areas already and could be easily ``plugged in'' or replaced in DogBot agent, for example a more complex visual system, i.e. a segmentation/detection neural network when using photo-realistic rendering, etc.

Last, there is no perfect formula to achieve good results. Part of it is the art of modeling the system and part is building insight from past failures and adapting, which itself is much the idea of (meta) reinforcement learning.

\bibliography{mybib}   

\begin{thebibliography}{13}
\providecommand{\natexlab}[1]{#1}
\providecommand{\url}[1]{\texttt{#1}}
\expandafter\ifx\csname urlstyle\endcsname\relax
  \providecommand{\doi}[1]{doi: #1}\else
  \providecommand{\doi}{doi: \begingroup \urlstyle{rm}\Url}\fi

\bibitem[Akkaya et~al.(2019)Akkaya, Andrychowicz, Chociej, Litwin, McGrew,
  Petron, Paino, Plappert, Powell, Ribas, et~al.]{akkaya2019solving}
I.~Akkaya, M.~Andrychowicz, M.~Chociej, M.~Litwin, B.~McGrew, A.~Petron,
  A.~Paino, M.~Plappert, G.~Powell, R.~Ribas, et~al.
\newblock Solving rubik's cube with a robot hand.
\newblock \emph{arXiv preprint arXiv:1910.07113}, 2019.

\bibitem[Barto and Mahadevan(2003)]{barto2003recent}
A.~G. Barto and S.~Mahadevan.
\newblock Recent advances in hierarchical reinforcement learning.
\newblock \emph{Discrete event dynamic systems}, 13\penalty0 (1-2):\penalty0
  41--77, 2003.

\bibitem[Bengio et~al.(2007)Bengio, LeCun, et~al.]{bengio2007scaling}
Y.~Bengio, Y.~LeCun, et~al.
\newblock Scaling learning algorithms towards ai.
\newblock \emph{Large-scale kernel machines}, 34\penalty0 (5):\penalty0 1--41,
  2007.

\bibitem[Bengio et~al.(2009)Bengio, Louradour, Collobert, and
  Weston]{bengio2009curriculum}
Y.~Bengio, J.~Louradour, R.~Collobert, and J.~Weston.
\newblock Curriculum learning.
\newblock In \emph{Proceedings of the 26th annual international conference on
  machine learning}, pages 41--48, 2009.

\bibitem[Juliani et~al.(2018)Juliani, Berges, Vckay, Gao, Henry, Mattar, and
  Lange]{juliani2018unity}
A.~Juliani, V.-P. Berges, E.~Vckay, Y.~Gao, H.~Henry, M.~Mattar, and D.~Lange.
\newblock Unity: A general platform for intelligent agents.
\newblock \emph{arXiv preprint arXiv:1809.02627}, 2018.

\bibitem[LeCun et~al.(2015)LeCun, Bengio, and Hinton]{lecun2015deep}
Y.~LeCun, Y.~Bengio, and G.~Hinton.
\newblock Deep learning.
\newblock \emph{nature}, 521\penalty0 (7553):\penalty0 436--444, 2015.

\bibitem[Lee et~al.(2019)Lee, Park, Lee, and Lee]{lee2019scalable}
S.~Lee, M.~Park, K.~Lee, and J.~Lee.
\newblock Scalable muscle-actuated human simulation and control.
\newblock \emph{ACM Transactions on Graphics (TOG)}, 38\penalty0 (4):\penalty0
  1--13, 2019.

\bibitem[Mnih et~al.(2013)Mnih, Kavukcuoglu, Silver, Graves, Antonoglou,
  Wierstra, and Riedmiller]{mnih2013playing}
V.~Mnih, K.~Kavukcuoglu, D.~Silver, A.~Graves, I.~Antonoglou, D.~Wierstra, and
  M.~Riedmiller.
\newblock Playing atari with deep reinforcement learning.
\newblock \emph{arXiv preprint arXiv:1312.5602}, 2013.

\bibitem[Schulman et~al.(2017)Schulman, Wolski, Dhariwal, Radford, and
  Klimov]{schulman2017proximal}
J.~Schulman, F.~Wolski, P.~Dhariwal, A.~Radford, and O.~Klimov.
\newblock Proximal policy optimization algorithms.
\newblock \emph{arXiv preprint arXiv:1707.06347}, 2017.

\bibitem[Silver et~al.(2017)Silver, Schrittwieser, Simonyan, Antonoglou, Huang,
  Guez, Hubert, Baker, Lai, Bolton, et~al.]{silver2017mastering}
D.~Silver, J.~Schrittwieser, K.~Simonyan, I.~Antonoglou, A.~Huang, A.~Guez,
  T.~Hubert, L.~Baker, M.~Lai, A.~Bolton, et~al.
\newblock Mastering the game of go without human knowledge.
\newblock \emph{Nature}, 550\penalty0 (7676):\penalty0 354--359, 2017.

\bibitem[Sutton and Barto(2018)]{sutton2018reinforcement}
R.~S. Sutton and A.~G. Barto.
\newblock \emph{Reinforcement learning: An introduction}.
\newblock MIT press, 2018.

\bibitem[Unity3D(2018)]{puppo}
Unity3D.
\newblock Puppo, the corgi: Cuteness overload with the unity ml-agents toolkit,
  2018.
\newblock URL
  \url{https://blogs.unity3d.com/pt/2018/10/02/puppo-the-corgi-cuteness-overload-with-the-unity-ml-agents-toolkit/}.

\bibitem[Vinyals et~al.(2019)Vinyals, Babuschkin, Czarnecki, Mathieu, Dudzik,
  Chung, Choi, Powell, Ewalds, Georgiev, et~al.]{vinyals2019grandmaster}
O.~Vinyals, I.~Babuschkin, W.~M. Czarnecki, M.~Mathieu, A.~Dudzik, J.~Chung,
  D.~H. Choi, R.~Powell, T.~Ewalds, P.~Georgiev, et~al.
\newblock Grandmaster level in starcraft ii using multi-agent reinforcement
  learning.
\newblock \emph{Nature}, 575\penalty0 (7782):\penalty0 350--354, 2019.

\end{thebibliography}
\end{document}